\documentclass{article}

\usepackage[preprint]{neurips_2022}

\usepackage[utf8]{inputenc} 
\usepackage[T1]{fontenc}    
\usepackage{hyperref}       
\usepackage{url}            
\usepackage{booktabs}       
\usepackage{amsfonts}       
\usepackage{nicefrac}       
\usepackage{microtype}      
\usepackage{xcolor}         
\usepackage{graphicx}       
\usepackage{float}          
\usepackage{longtable}      
\usepackage{array}          
\usepackage{enumitem}       

\usepackage[table]{xcolor}

\title{Visual Reasoning Benchmark: Evaluating Multimodal LLMs on Classroom-Authentic Visual Problems from Primary Education}

\author{
  \textbf{Mohamed Huti$^{1}$ \quad Alasdair Mackintosh$^{1}$ \quad Amy Waldock$^{1}$ \quad Dominic Andrews$^{1}$} \\[10pt]
  \textbf{Maxime Leli\`evre$^{1}$ \quad Moritz Boos$^{1}$ \quad Tobias Murray$^{1}$} \\[10pt]
  \textbf{Paul Atherton$^{1}$ \quad Robin A. A. Ince$^{1}$ \quad Oliver G. B. Garrod$^{1}$} \\[20pt]
  {\normalsize $^1$Fab AI}
}

\begin{document}

\maketitle

\begin{abstract}
AI models have achieved state-of-the-art results in textual reasoning; however, their ability to reason over spatial and relational structures remains a critical bottleneck---particularly in early-grade maths, which relies heavily on visuals. This paper introduces the visual reasoning benchmark (VRB), a novel dataset designed to evaluate Multimodal Large Language Models (MLLMs) on their ability to solve authentic visual problems from classrooms. This benchmark is built on a set of 701 questions sourced from primary school examinations in Zambia and India, which cover a range of tasks such as reasoning by analogy, pattern completion, and spatial matching. We outline the methodology and development of the benchmark which intentionally uses unedited, minimal-text images to test if models can meet realistic needs of primary education. Our findings reveal a ``jagged frontier'' of capability where models demonstrate better proficiency in static skills such as counting and scaling, but reach a distinct ``spatial ceiling'' when faced with dynamic operations like folding, reflection, and rotation. These weaknesses pose a risk for classroom use on visual reasoning problems, with the potential for incorrect marking, false scaffolding, and reinforcing student misconceptions. Consequently, education-focused benchmarks like the VRB are essential for determining the functional boundaries of multimodal tools used in classrooms.
\end{abstract}

\section{Introduction}

AI models have achieved state-of-the-art progress on a wide range of reasoning tasks, from natural language understanding and knowledge retrieval to symbolic mathematics. Benchmarks such as GSM8K \citep{cobbe2021gsm8k}, MATH \citep{hendrycks2021math} and MMLU \citep{hendrycks2021mmlu} now demonstrate above-human performance when reasoning is presented in purely textual form. Yet these advances do not extend to a core dimension of intelligence: the ability to reason visually through patterns, diagrams, and spatial relationships \citep{wang2024spatial}.
Classic measures such as Raven's Progressive Matrices \citep{raven1936} intentionally minimise language to assess abstract reasoning beyond vocabulary or cultural familiarity, underscoring the importance of non-verbal inference in cognition and assessments. In educational practice, such assessments have been used in school entry decisions, in identifying gifted learners and in the equitable evaluation of children with dyslexia or those learning in a language other than their mother tongue.

Visual reasoning is particularly important in learning foundational mathematics. Effective mathematical problem solving commonly involves constructing and manipulating appropriate visual representations (e.g., diagrams, number lines, arrays) to support inference \citep{purcar2024}. Evidence from early primary classrooms indicates that providing pupils with systematic opportunities to develop such representations enhances arithmetic problem-solving performance \citep{purcar2024}, and that instruction targeting spatial skills contributes to more advanced computational reasoning \citep{parkinson2025}. Maths curricula usually utilise concrete-pictorial-abstract learning pathways, where various visual representations are used to support learning by providing connections between children's experience and abstract conceptions.
Consequently, to be effective in supporting student learning in the early years of school, AI models need to be capable of understanding visual reasoning problems well. For example, an AI tutor must be able to recognize student errors in classroom visual problems and provide correct explanations, and for accurate automatic grading, the model needs high accuracy in problem recognition and answering.

Despite rapid progress, contemporary Multimodal Large Language Models (MLLMs) often falter when the decisive information is visual. Recent studies across multimodal and abstract reasoning benchmarks reveal a persistent performance gap between textual and visual modalities, suggesting that models can leverage linguistic cues effectively but still struggle to perceive and reason over the spatial and relational structures encoded in images \citep{xu2025visulogic}.

These findings raise a practical question for education: To what extent do current models possess the kind of visual reasoning capabilities needed to be useful in classrooms? Rather than measuring abstract reasoning ability for complex problems that many adults would find challenging, we focus on whether models can handle the kind of visual problems that primary students would encounter, thus establishing a minimum capability threshold for models to support learners and teachers effectively in visual problem-solving contexts.

The Visual Reasoning Benchmark (VRB) introduced in this paper addresses this question by emphasising minimal-text, classroom-authentic visual problems. In education, visual reasoning (also called non-verbal reasoning) is defined as ``using pictures, images or diagrams effectively for solving tasks of higher-order thinking'' \citep{natsheh2014}. Here, we use visual reasoning to mean \emph{deriving the correct answer by perceiving and operating on spatial and relational structure present in an image}.
The VRB design complements multimodal datasets such as MathVista \citep{xie2023mathvista} and Math-V \citep{lu2024mathv} which include visual problems but retain textual statements and choices, while aligning more closely with VisuLogic \citep{xu2025visulogic} which deliberately targets non-verbal reasoning. Our contribution is to bring this challenge into authentic educational contexts. By drawing directly on primary-level assessment tasks from Zambia and India, VRB provides the first large-scale, classroom-grounded benchmark for evaluating visual reasoning in low- and middle-income country (LMIC) settings. It uses unedited questions, including slight issues from photocopying or production common in an LMIC setting, revealing how a model might handle genuine student questions (for example, those presented to a chatbot). This is an essential step to determining when such systems could become genuinely useful for learning support. We estimate the minimum capability threshold for classroom usefulness at 94\%, the proportion of questions on which three adult annotators independently agreed on the answer (see Section~\ref{sec:human_review}). In this way, the benchmark is designed to test if models meet grade-level expectations in visual reasoning and to indicate their suitability for use by teachers and students in lesson support and design.

\section{Related Work}

Research on visual reasoning has developed along distinct trajectories, each addressing different aspects of the challenge.

\subsection{Synthetic Abstract Reasoning}

The Abstraction and Reasoning Corpus (ARC) \citep{chollet2019} represents the purest test of abstract visual reasoning, presenting grid-based challenges where participants must uncover hidden transformation rules such as symmetry or repetition. Humans achieve 98--100\% accuracy, but frontier models reach 85--95\% (ARC-AGI-1) and around 70\% (ARC-AGI-2) \citep{chollet2025}. MARVEL \citep{jiang2024marvel} decomposes this challenge into perception and abstraction components, by testing six core patterns such as 3D-geometry and temporal movement across different visual problems. It reveals that despite high symbolic logic scores, MLLMs often perform near chance specifically on non-verbal reasoning tasks because they struggle with a persistent perceptual bottleneck.

\subsection{Academic Mathematical Reasoning}

A substantial body of work targets mathematical problems based on the interpretation of diagrams, charts, or figures. MathVista \citep{xie2023mathvista} compiles 6,141 problems from 28 multimodal datasets spanning geometry, algebra, and scientific plots. While frontier models at the time of launch like Gemini 2.0 Flash achieved 73.1\%---surpassing the 60.3\% human baseline---this success occurs primarily in settings with high textual scaffolding where linguistic cues provide a redundant path to the solution. MATH-V \citep{lu2024mathv} extends this approach with 3,040 diagram-augmented contest problems across 16 topics and five difficulty levels, explicitly ensuring visual elements are non-redundant and integral to problem-solving. MathVerse \citep{zhang2024mathverse} provides the most systematic analysis by using 15,000 questions which include text and diagrams, where some answers depend more on information in the text and others on the diagrams. Their findings reveal a consistent pattern: accuracy declines for questions where the critical information lies in the diagrams. GPT-4V performance drops from 54.7\% (Text-Dominant) to 31.6\% (Vision-Only), and several models actually improve when images are removed entirely. MV-MATH \citep{wang2025mvmath} raises the bar with 2,009 problems requiring reasoning across 2--5 images per question, where the best models achieve 34\% compared to 75--80\% for humans. Across these datasets, multimodal models exploit textual scaffolding effectively, but their diagram-grounded reasoning remains brittle.

\subsection{Research-Focused Visual Reasoning}

Recent work has moved toward minimising textual scaffolding to isolate genuine visual reasoning capabilities. VisuLogic \citep{xu2025visulogic} introduces 1,000 fully non-verbal, human-verified puzzles across six categories, with humans scoring 51.4\% while leading MLLMs remain below 30\%. MM-IQ \citep{cai2025mmiq} extends this approach to a larger scale (2,700 test items) across eight reasoning paradigms, yielding similar outcomes (33\% for models vs 51\% for humans). BabyVision \citep{chen2026babyvision} targets an even more fundamental level, testing core visual abilities that humans acquire before language. Across 388 items spanning four categories, the best-performing MLLM scores 49.7\%, falling below the level of six-year-old children and far short of the adult baseline (94.1\%). VERIFY \citep{liu2025verify} emphasises fidelity by pairing 600 diagrammatic items with human-annotated reasoning steps. Top models achieve 21.7\% accuracy and frequently generate correct answers with flawed reasoning. ME2 \citep{park2025me2} shifts focus from accuracy to explanatory fidelity, requiring solution steps that reference diagram annotations; even strong models struggle to align explanations with visuals. These efforts mark a shift from outcome-only scoring toward process-aware evaluation, revealing that models frequently show incorrect reasoning even when answers are correct.

\subsection{The Educational Authenticity Gap}

Collectively, these benchmarks reveal a consistent pattern: models that excel at text-based reasoning often collapse when equivalent reasoning must be conducted visually. This reflects dual challenges of perception (detecting and segmenting visual elements) and abstraction (inferring relational rules and generalizing across contexts). While contemporary vision encoders provide strong perceptual grounding, abstraction remains elusive.

In educational practice, another critical gap emerges in this landscape: \textbf{none of these core visual reasoning benchmarks are grounded in authentic primary-education assessment contexts}. Synthetic puzzles like ARC or MM-IQ probe abstraction in isolation but lack pedagogical relevance. Contest-style datasets like MATH-V target advanced mathematical reasoning but represent specialised academic contexts rather than foundational learning. Research-focused benchmarks often yield low-to-moderate human performance, suggesting a disconnection from natural learning tasks and classroom relevance.

Our benchmark addresses this gap by drawing directly from primary-level educational assessments used in real classrooms. We focus on visual reasoning tasks that millions of children successfully navigate as part of their learning journeys. This approach not only provides a real-world test of reasoning capabilities but also establishes clear expectations based on age-appropriate educational standards. In doing so, we create the first large-scale evaluation that is simultaneously rigorous for AI systems and directly relevant to educational contexts where such reasoning is integral to effective learning.

\section{Methodology}

\subsection{Source of Questions}

We extracted multiple-choice questions (MCQs) from Zambia's National End of Primary Exams Special Paper 2 (Non-Verbal Reasoning) and the Jawahar Navodaya Vidyalaya's Selection Test Class 6 (JNVST Class 6) from India.

Special Paper 2 from Zambia is administered as part of Grade 7's End of Primary Leaving Exams. Together with Special Paper 1 (verbal reasoning), it serves as an aptitude test. The aim of the test is to assess cognitive skills beyond curriculum subject areas and to provide information to support secondary school enrolment. We sourced questions from Special Paper 2 Exam Papers administered in 2018, 2019, 2021, and 2022.

JNVST is the entrance exam for the Jawahar Navodaya Vidyalayas (JNV) schools, a network of co-educational, residential schools fully financed and administered by the Government of India. It is a national-level examination based on exams designed by the Central Board of Secondary Education (CBSE) and conducted independently in each state by the Navodaya Vidyalaya Samiti, an autonomous organisation\footnote{\url{https://examcart.in/community/blogs/jnvst-class-6-exam-date-syllabus-eligibility-pattern-pyqs}}. We sourced questions from JNVST administered in 2014--2020, 2022, and 2024.

\subsection{Question Processing}

\subsubsection*{Question Extraction and Pre-processing}

The MCQ questions were extracted from the exam PDFs. Items were extracted at the question level by manual region-of-interest cropping (figures + options) using a PDF viewer (Preview) with occasional readjustment. Crops were exported as black-and-white JPEGs using default viewer settings. Note that all the original source questions are also black-and-white prints. To reflect how a model might be used in practice by students and teachers, we applied no normalisation or sharpening. We did not remove on-page text (e.g.\ option letters) if they were a part of the cropped region. Each MCQ received a deterministic Question ID keyed to its source. A lightweight index links Question ID to image path and the source's question.
Models were evaluated in an image-first, minimal-text method---mirroring recent research that minimises linguistic scaffolding to isolate vision-centric reasoning. Prompts standardise response format only; they never include hints, examples, or domain content.
Some of the exams also had linked answers, which were provided in separate documents. These were parsed and matched to the extracted questions, although additional verification took place as described in the next section.
To prevent benchmark leakage, we release all evaluation prompts and code needed to reproduce our pipeline, but we do not publicly distribute the question images or answer keys. We will work directly with model developers to facilitate evaluation of new models against the benchmark.

\subsection{Human Review, Marking and Annotation}
\label{sec:human_review}

To ensure the quality of the MCQs and for selection and annotation, each MCQ was independently checked, answered, and categorised by three human markers (two education experts and one non-expert).
The markers' answers were used to provide the verified answer which models could be scored against. Where markers felt there was no correct answer, they recorded a `refuse' rather than a random selection. In order to be retained in the dataset, we selected those MCQs with an answer with a majority consensus between the three human markers. 34 questions were excluded as there was no majority consensus answer or 2 or more markers gave `refuse'. A further 10 questions were removed from the benchmark due to fatal errors (9 with corrupted or incorrect images and 1 with a question presentation error).

In total, 658 questions were retained with full consensus answers across all markers, and an additional 43 MCQs where 2 out of 3 markers agreed. We chose to retain these questions with a 2 out of 3 answer consensus, because there was still a clear answer, but these were likely to be more challenging as one human had failed to answer correctly. Our aim was to ensure the benchmark would test even the most capable models rigorously. This gave a final dataset containing 701 questions.

\begin{figure}[H]
    \centering
    \includegraphics[width=0.85\textwidth]{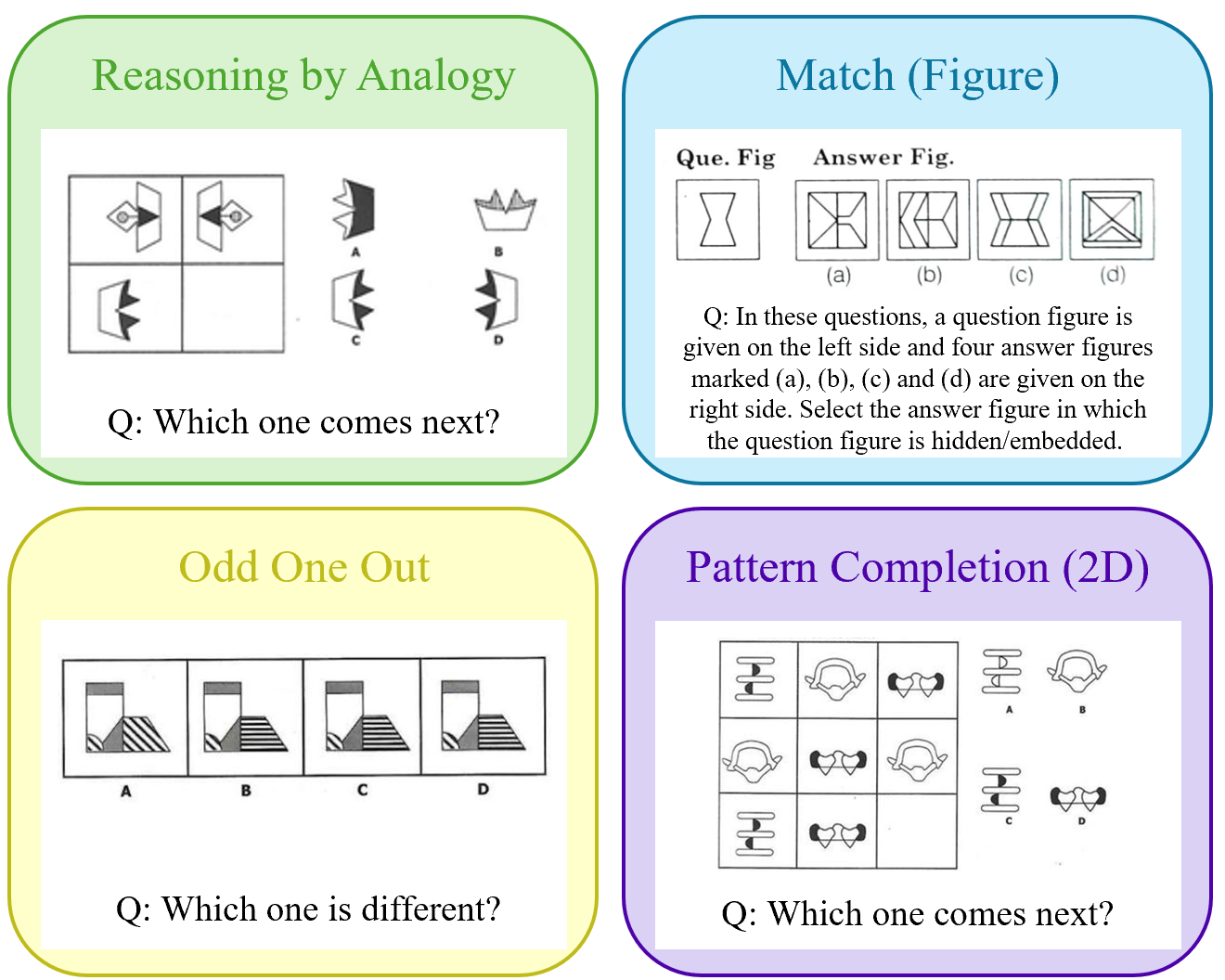}
    \includegraphics[width=0.85\textwidth]{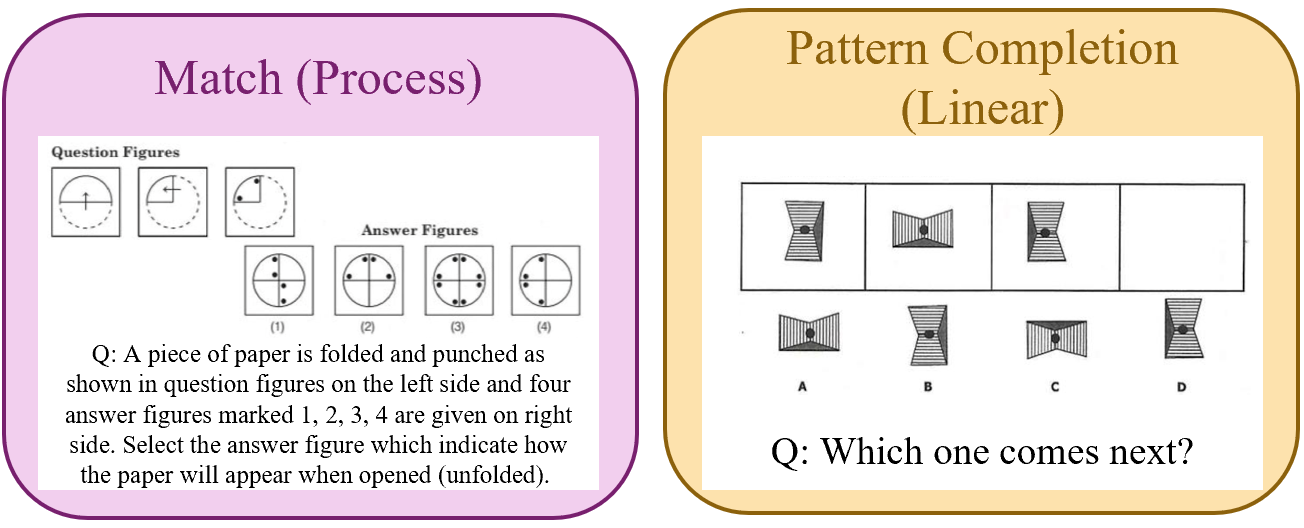}
    \caption{Question examples showing the six task categories in our Visual Reasoning Benchmark.}
    \label{fig:task_examples}
\end{figure}

\begin{figure}[H]
    \centering
    \includegraphics[width=0.85\textwidth]{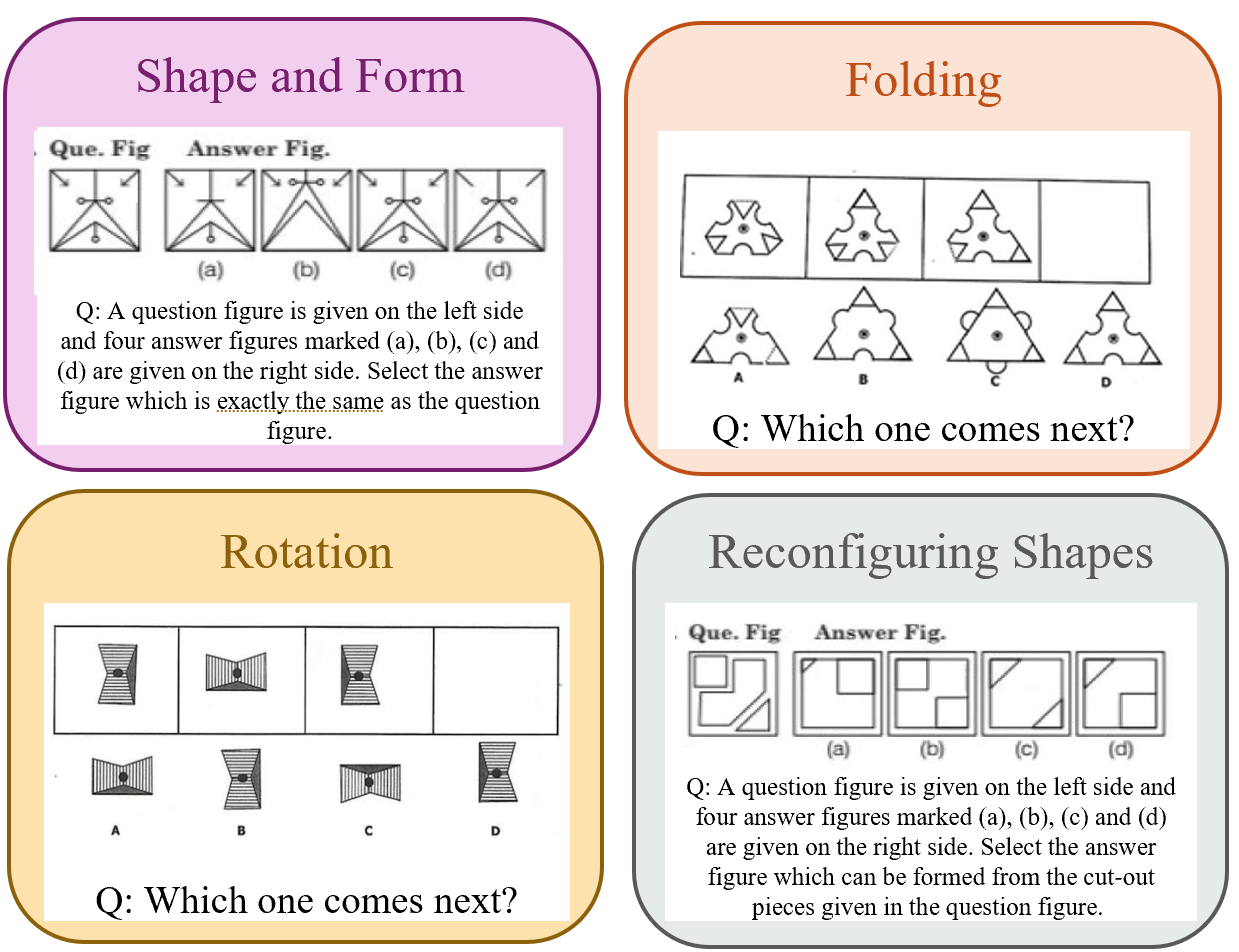}
    \includegraphics[width=0.85\textwidth]{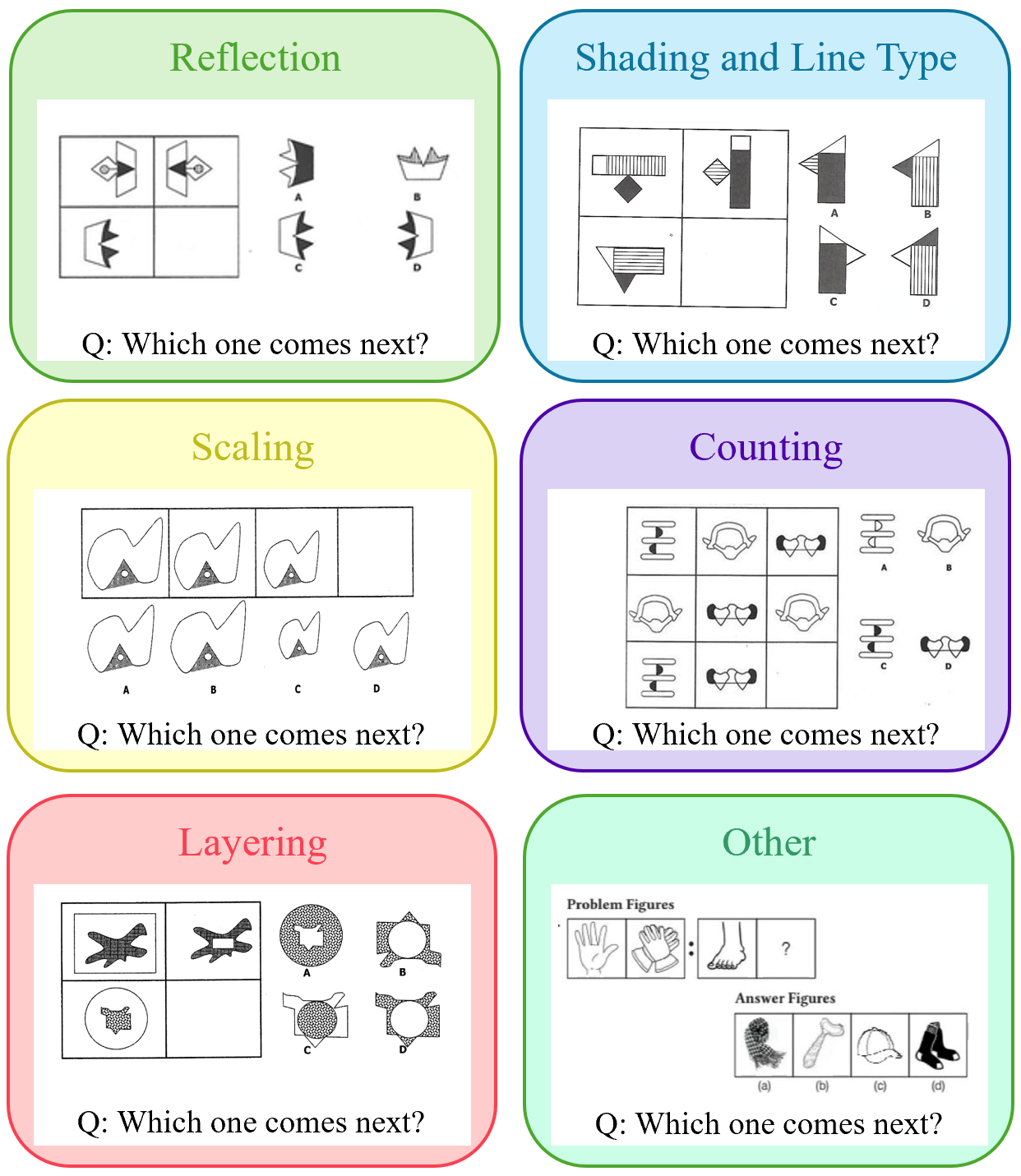}
    \caption{Question examples showing the ten skill tags in our Visual Reasoning Benchmark.}
    \label{fig:skill_examples}
\end{figure}

\begin{figure}[H]
    \centering
    \includegraphics[width=0.75\textwidth]{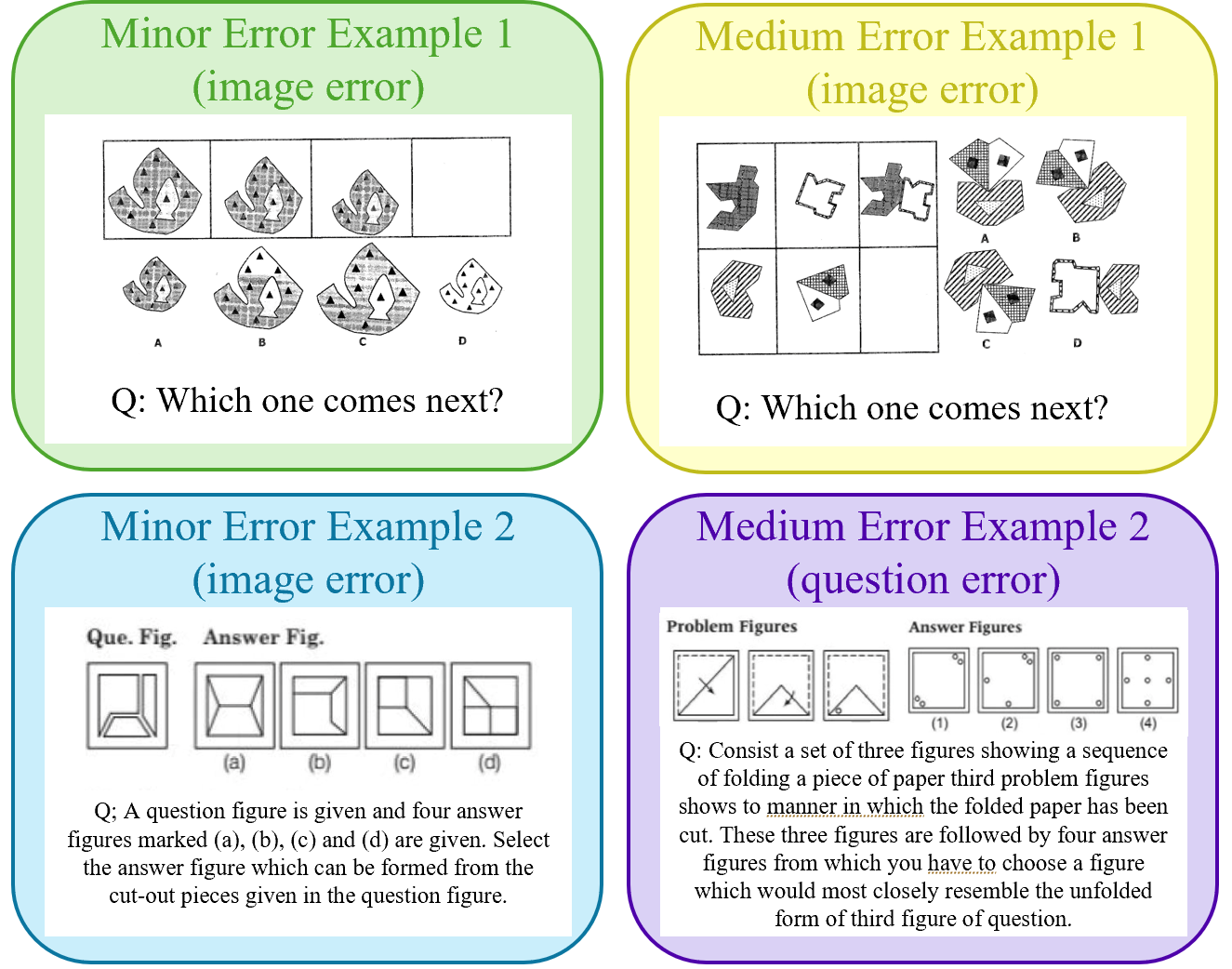}
    \caption{Question examples showing the two types of error in our Visual Reasoning Benchmark.}
    \label{fig:error_examples}
\end{figure}

To support analysis of this final dataset, we annotated each question by `task', and by `skill'. The task considers how the question was structured (e.g.\ odd one out, pattern completion, matching). The skill considers the underlying visual operation(s) (e.g.\ folding, counting, rotating) needed to solve the questions.
To identify task and skill category names, a review was done of sources of visual reasoning questions. The categories we selected were mostly derived from guidance material used to support the UK `11+' Secondary Entrance Exams \citep{broadbent2018}. Whilst the `11+' does not have official question categories, this guidance material (for a well-recognised exam taken at the same age as our question set) offered a useful starting point. Where the existing categories did not match our question set well, we adapted. For skills, we added Scaling and also combined two others to create Reconfiguring Shapes and Layering. For tasks, we added Match (process) and Match (figure). In total, we used 6 tasks and 10 skills. See Figures~\ref{fig:task_examples},~\ref{fig:skill_examples} and~\ref{fig:error_examples} for question samples and Appendix~\ref{app:definitions} for definitions of categories and tags.

For tasks, only one label was assigned. All task labels had 2 out of 3 or full marker agreement, with the majority label assigned used in the final annotations.
For skills, flexibility was given to assign more than one skill tag because some questions involved observation of two or more metrics, for example, a shape being both rotated and its shading changing. Similarly, the final skill tags were assigned where there was majority agreement between markers (i.e.\ if 2 out of 3 markers or more assigned a tag, it was used in the final annotations).

Where appropriate and necessary, a pedagogy expert made minor edits to MCQs. Changes focused on removing irrelevant information, such as instructions for how to fill in answer booklets. However, MCQs were intentionally left as close to the original as possible to determine how models performed with genuine student questions.

Finally, to evaluate the robustness of multimodal models, we also annotated the quality of the question image. We categorised items based on their potential impact on student comprehension: No Error, Minor (Slight artefacts), and Moderate (clear visual noise that requires more cognitive effort to interpret). (See Appendix~\ref{app:errors} for full definitions).

\section{Results}

\subsection{Visual Reasoning Benchmark (VRB) Performance}

\begin{figure}[H]
    \centering
    \includegraphics[width=\textwidth]{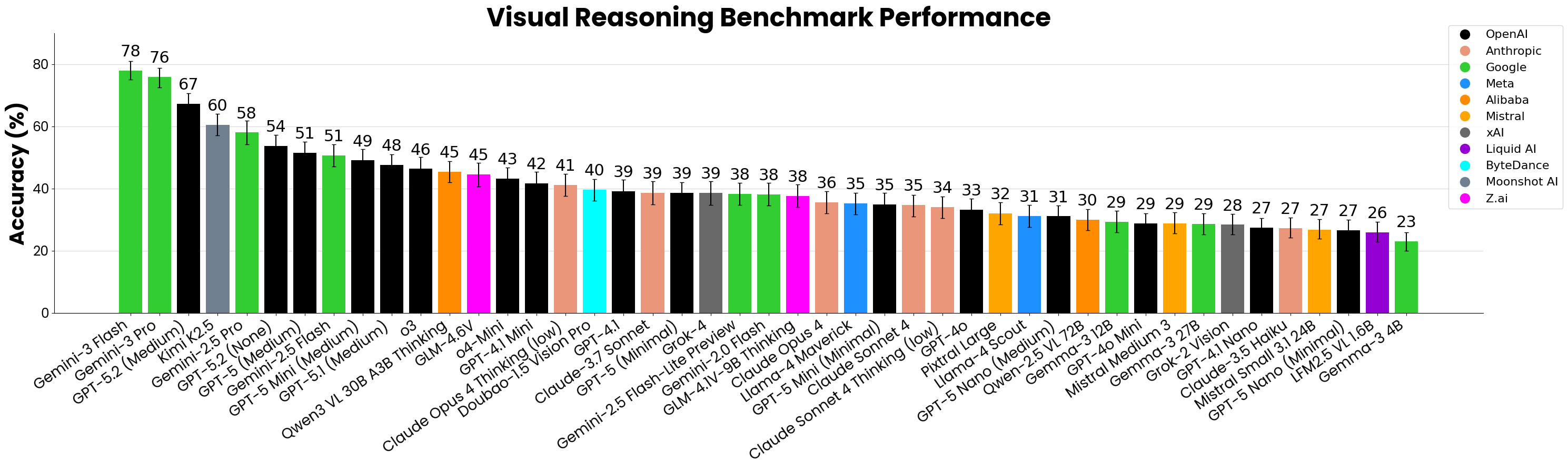}
    \caption{Accuracy on VRB for a subset of models. Errors show 95\% bootstrap confidence intervals.}
    \label{fig:vrb_accuracy}
\end{figure}

At the time of writing, 45 multimodal models have been evaluated covering both proprietary and open weighted models from major providers and spanning a wide range of sizes and price points. All models answer 4-option multiple-choice questions using a single image, minimal-text prompt. We report item-level accuracy as the primary metric, together with 95\% bootstrap confidence intervals over 1000 resamples (Figure~\ref{fig:vrb_accuracy}). Chance performance on VRB is 25\%, so scores near this level are equivalent to random guessing over the options.

Across the full panel, accuracies range from 23\% up to 78\%. Gemini-3.0 Flash and Gemini-3.0 Pro attain the highest scores at 78\% and 76\% respectively. However, given that these questions are targeted at end-of-primary students and do not require advanced subject knowledge, this means that even the currently highest scores are relatively weak, and suggests the benchmark is currently far from saturation. This is echoed by the overall wide variation, and the long tail of models close to the 25\% chance level, highlighting the need for improvements in this area by model developers.

Taken together, these results show that primary-level visual reasoning remains challenging for current multimodal LLMs. Even the best available models only solve a subset of items reliably and performance varies substantially across the model landscape. In the following sections, we unpack this variation by model type, reasoning capability, cost and later by skill profiles.

\subsubsection{Model groups: weights and reasoning capability}

\begin{figure}[H]
    \centering
    \includegraphics[width=\textwidth]{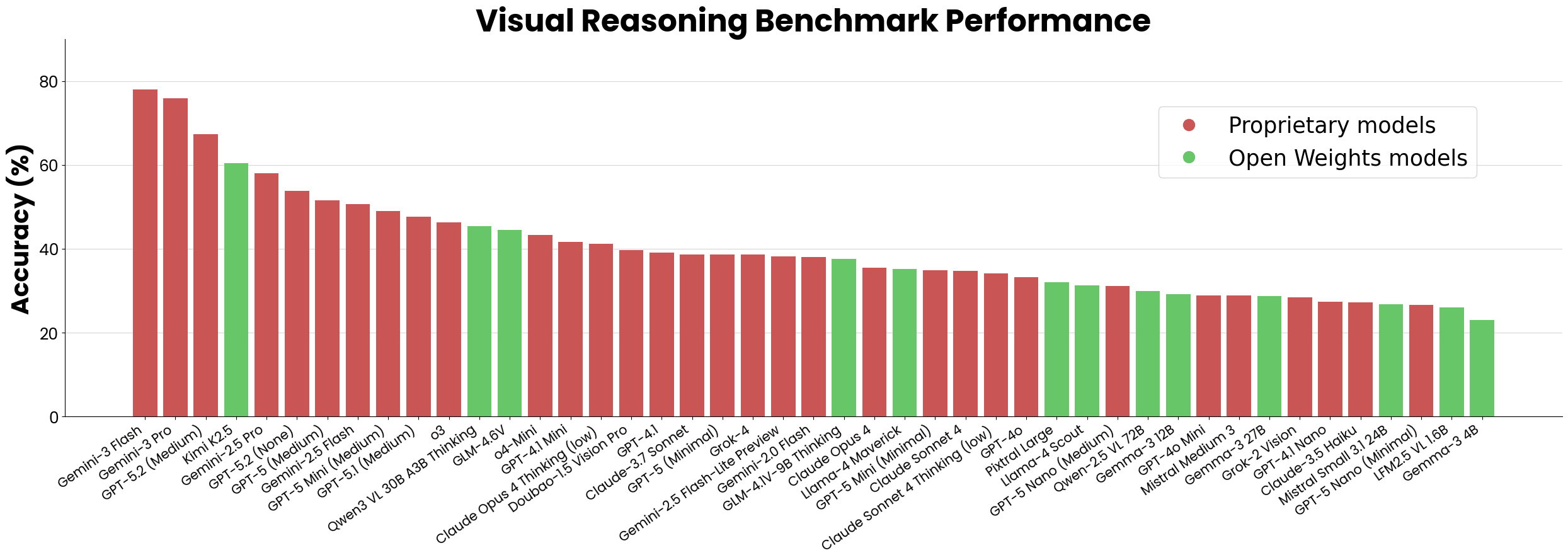}
    \caption{Accuracy on the Visual Reasoning Benchmark (VRB) by Weights Availability.}
    \label{fig:weights}
\end{figure}

Figure~\ref{fig:weights} compares performance on VRB for proprietary (closed-weight, API-access) models and open-weight models. Proprietary models largely dominate the top of the leaderboard; Kimi-K2.5 is the main exception at 60\% and then GLM-4.6V and QWEN3-VL are the only other open-weight models appearing in the upper half, reaching an accuracy $\approx$45\%. Small open models such as Gemma-3 4B and Mistral Small 2.1 3B remain around the mid 20\% range effectively at or near the chance baseline. It is also worth noting that many of the strongest open-weight reasoning models currently available are text-only and therefore do not appear in this multimodal evaluation.

These results indicate that, as of January 2026, open-weight models mostly lag behind proprietary systems in visual reasoning on VRB, though the 60\% performance of Kimi-K2.5 represents a significant step forward in closing this gap. However, in LMIC contexts where on-device or locally hosted models are preferred for privacy, connectivity, and cost---this progress meets a practical limit. Because Kimi-K2.5 is a trillion-parameter model requiring massive hardware resources, it remains unsuitable for low resource local hosting. Consequently, our results highlight a sharp trade-off: the scale required for open-weight models to achieve frontier-level visual reasoning creates an infrastructure barrier that negates the cost and accessibility benefits of local hosting.

\begin{figure}[H]
    \centering
    \includegraphics[width=\textwidth]{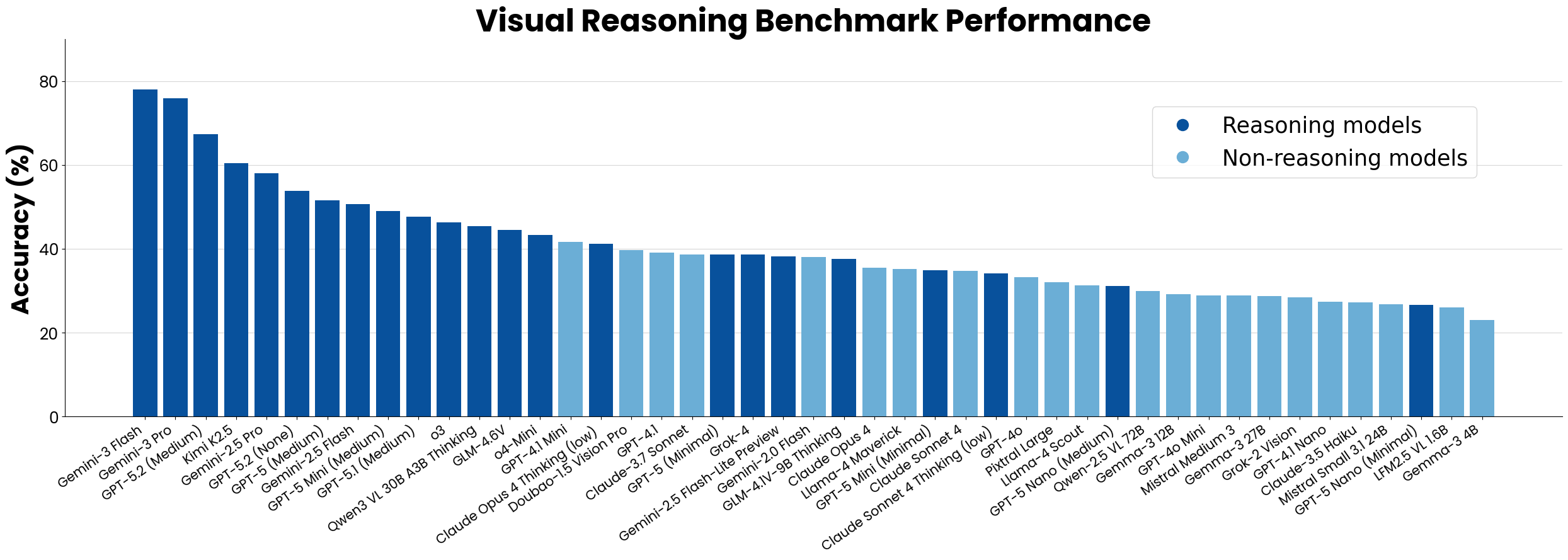}
    \caption{Accuracy on the Visual Reasoning Benchmark (VRB) by Reasoning Capability.}
    \label{fig:reasoning}
\end{figure}

To examine the role of explicit reasoning, we group models into reasoning variants and non-reasoning variants (Figure~\ref{fig:reasoning}). Reasoning models are those that expose dedicated ``thinking'' or chain-of-thought modes (e.g.\ Gemini 3.0, GPT-5 (medium), Claude Opus 4 Thinking), while their corresponding base chat models and other standard instruction-following models are treated as non-reasoning variants.

\subsubsection{Accuracy vs Cost: The Visual Value Frontier}

\begin{figure}[H]
    \centering
    \includegraphics[width=\textwidth]{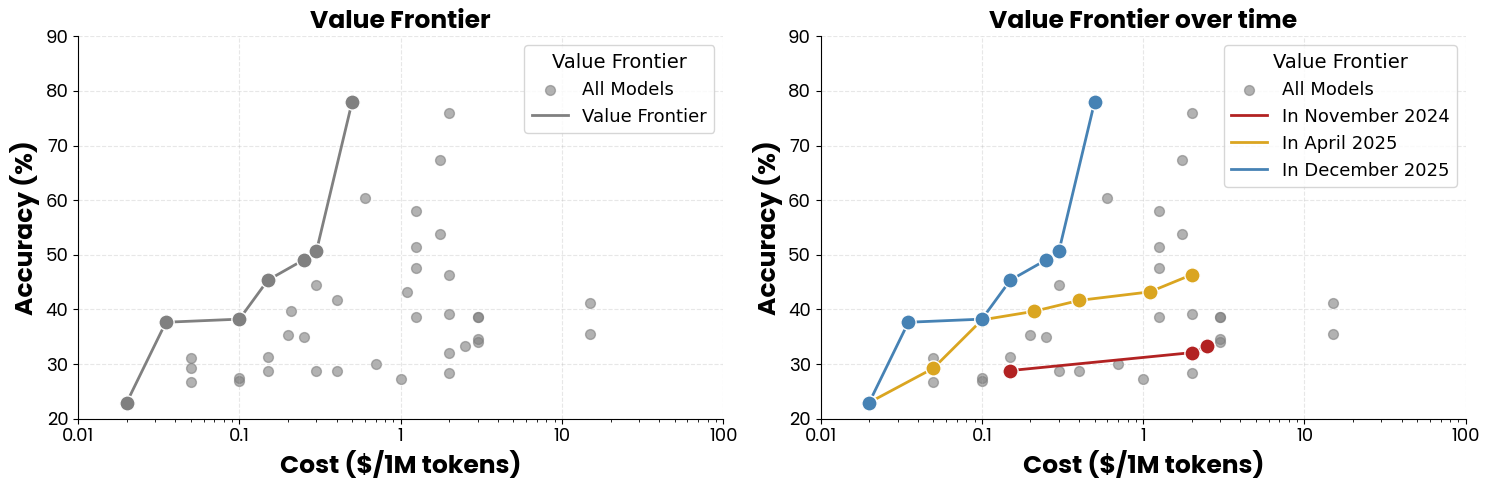}
    \caption{Left: \textbf{Cost vs accuracy,} the best models being on the value frontier (grey line). Right: \textbf{Value Frontier over time} on VRB.}
    \label{fig:cost}
\end{figure}

For models with published pricing, we record the input token cost in USD per million tokens and examine the trade-off between VRB accuracy and inference cost (Figure~\ref{fig:cost}). Token prices span nearly four orders of magnitude, from around \textbf{\$0.01} to close to \textbf{\$100} per million input tokens. At the very low end of this range, models achieve only $\approx$\textbf{23\%} accuracy and are unlikely to be useful for most classroom tasks involving visual reasoning. Around \textbf{\$0.10--\$0.20} per million tokens, the best models reach accuracies of roughly \textbf{38--40\%}. The current value frontier is dominated at the high end by a notable anomaly: Gemini-3 Flash, achieving an accuracy of 78\% at a price point of only \textbf{\$0.50}, exceeding the performance of models that cost ten times as much.

For any given cost band there is typically a spread of around 15--20 percentage points between the lowest- and highest-performing models, indicating substantial variation in ``visual value for money''.

We also track how this value frontier evolves over time by comparing snapshots from late 2024, mid-2025 and late 2025. Across almost all price points, the frontier shifts upwards: for example, at around \textbf{\$0.10/M} the best available accuracy increases from $\approx$30\% to $\approx$38--39\% over this period, while at higher costs the top models move from the low-30s to around 60\%. This pattern mirrors trends in text-based benchmarks, but here it shows that non-verbal visual reasoning capabilities are also improving rapidly. VRB is thus sensitive enough to register incremental gains in visual reasoning and to highlight models that offer the best trade-off between performance, openness and cost.

\subsection{Task and Skill Level Performance}

We analysed performance across two dimensions to investigate which types of questions MLLMs are performing better on, and where improvements are most needed. As highlighted in Section~\ref{sec:human_review}\footnote{See Figures~\ref{fig:task_examples},~\ref{fig:skill_examples} and~\ref{fig:error_examples} for question samples and Appendix~\ref{app:definitions} for definitions of categories and tags.}:

\begin{itemize}
    \item \textbf{Task} considers how the question was structured (e.g.\ odd one out, pattern completion, matching).
    \item \textbf{Skill} considers the underlying visual operation(s) (e.g.\ folding, counting, rotating) needed to solve the questions.
\end{itemize}

These dimensions help us to understand and distinguish between where the models fail (the visual layout) and why they fail (the cognitive bottleneck).

\subsubsection{Task formats and performance}

\begin{figure}[H]
    \centering
    \includegraphics[width=\textwidth]{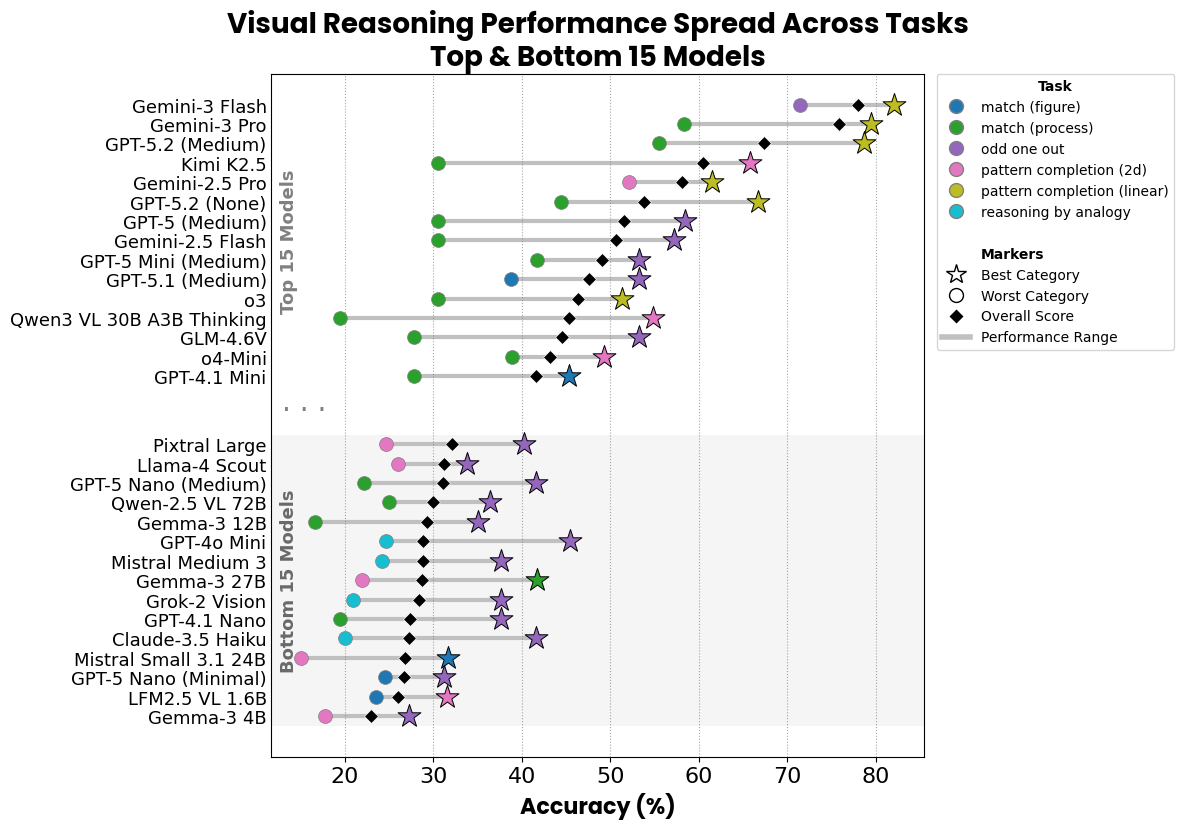}
    \caption{The spread of performance over tasks for the top and bottom 15 models.}
    \label{fig:task_performance}
\end{figure}

To uncover the mask of overall performance accuracy we analysed the accuracy by looking at the task format. Figure~\ref{fig:task_performance} shows the performance spread across task categories for the top and bottom 15 models, highlighting each model's overall score (diamond), best task (star), worst task (circle), and the range between the best and the worst tasks (grey line).

For the Top 15 models, we can see that there is a large volatility within models. While many top-tier models show a 10--25 percentage point gap between their best and worst task, this gap is even wider in newer open-source models like Kimi-K2.5 and Qwen3-VL, where the difference is nearly 35 points. We observe that even as open-source models become more capable on average, their performance becomes more inconsistent across different types of visual logic. The strongest models achieve the highest accuracies on \textit{pattern completion (linear)} tasks which suggests that models find it easier to extend sequences once a rule is inferred. On the other hand, the worst-performing task for most frontier models is \textit{Match (process)}, which could be due to the multi-step transformations required rather than a simple visual comparison.

In the bottom 15 models, overall accuracy clusters much lower, around the 20\%--40\% range. These models perform better on \textit{Odd One Out} tasks, which focus on direct comparisons rather than learning and applying a transformational rule.

Task formats give a useful view on which question types models can struggle on. However, a single task category can use multiple visual operations e.g.\ \textit{Match (Figure)} might only require \textit{counting} in one question and \textit{rotation} or \textit{folding} in another. Because skill tags are unevenly distributed across tasks and can co-occur within the same item, task-level results do not show us the full picture. The next section shifts from presentation to the source of difficulty by estimating how each visual skill affects accuracy.

\subsubsection{Skill effects on question performance}

\begin{figure}[H]
    \centering
    \includegraphics[width=\textwidth]{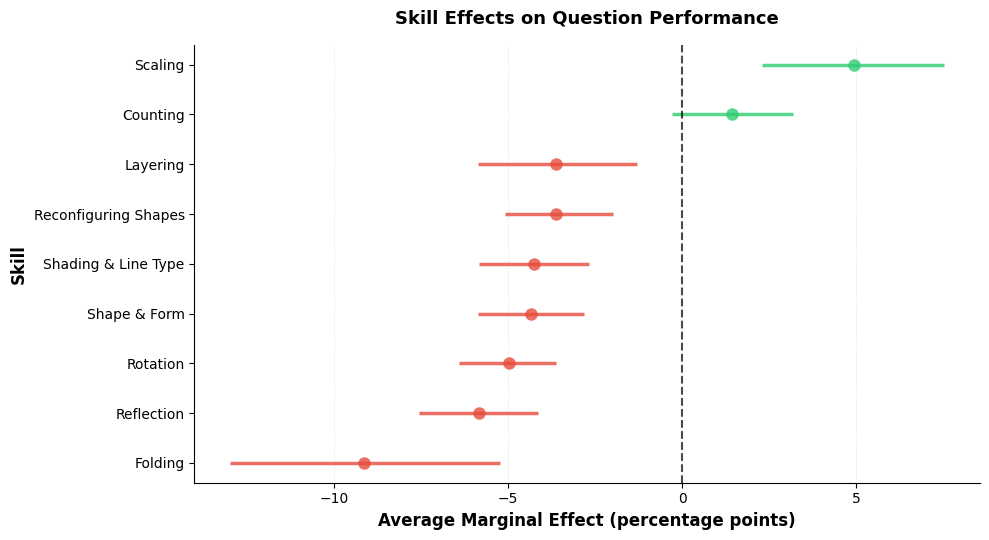}
    \caption{\textbf{Average marginal effects of specific skills on question performance.} This plot shows the predicted change in accuracy (percentage points) associated with each skill tag, accounting for model identity, dataset, and task category. Uncertainty is represented by 95\% bootstrap confidence intervals.}
    \label{fig:skill_effects}
\end{figure}

To find out which visual operations AI models are finding the most challenging, we calculated the marginal effect of each skill on item performance while controlling for model, dataset, and task. The resulting marginal effects represent the average change in accuracy (percentage points) that is associated with the presence of a specific skill.

There are two skills that stand out to facilitate correct responses. Items that are tagged with \textit{Counting} and \textit{Scaling} are strong predictors for success. \textit{Scaling} provides a boost of approximately +5 percentage points while \textit{Counting} has an increase of +1.4 points relative to the absence of these skills. This suggests that multimodal LLMs are relatively more proficient at detecting numerosity and recognising monotonic size changes.

All other skills correlate with a decrease in performance. \textit{Layering} and \textit{Reconfiguring Shapes} result in modest decreases of approximately $-3.5$ percentage points. The most significant drop happens in tasks requiring complex mental manipulation. \textit{Shading and Line Type} and \textit{Shape and Form} both lead to a decrease of $-4.5$ points, whilst \textit{Rotation} ($-5$ points) and \textit{Reflection} ($-5.8$ points) represent substantial hurdles. The most difficult skill is \textit{Folding} which is associated with a decrease of $-9$ points. These findings indicate that while models can ``see'' and ``count'' discrete elements, they fundamentally struggle with spatial transformations---the mental gymnastics of rotating, reflecting, or folding objects in 3D space.

\subsection{Error Analysis}

\begin{figure}[H]
    \centering
    \includegraphics[width=0.9\textwidth]{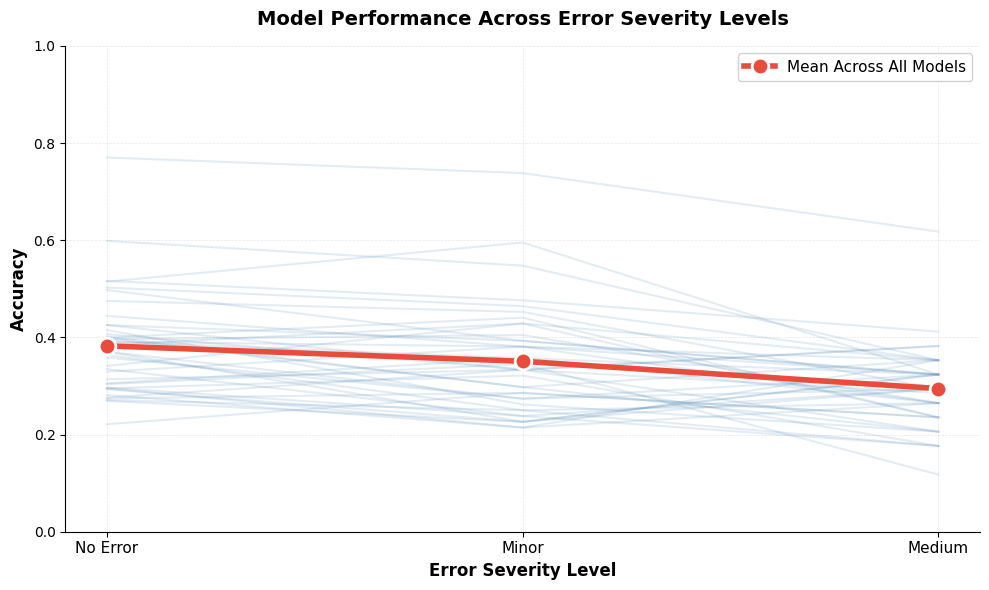}
    \caption{\textbf{Model accuracy across three error-severity levels} (No error, Minor, Medium). Thin lines show individual model trajectories; the thick red line shows the mean across all models.}
    \label{fig:error_analysis}
\end{figure}

A distinct contribution of VRB is the inclusion of classroom-authentic questions that contain noise such as photocopying artefacts (e.g., faded line segments and inconsistent shading) or minor production errors that are likely to be present in real-world use cases, particularly in LMIC settings.

As shown in Figure~\ref{fig:error_analysis}, we see a decline in overall accuracy as the visual quality decreases. The mean accuracy across all evaluated models drops from 38\% on pristine items to approximately 30\% on items with moderate error severity.

The steepest rate of decline was for the highest performing models. This suggests that high-performing frontier models achieve these scores through precise pixel matching---relying on low-level visual patterns that lack structural integrity, which is easily disrupted by the presence of artefacts. Lower-performing models nearer the chance threshold of 25\% experienced less variance due to the image quality, indicating that these architectures fail at extracting meaningful structural information regardless of the quality of the input.

\section{Discussion}

VRB was designed to investigate a practical question: Are current multimodal LLMs equipped to navigate the non-verbal reasoning challenges that are typically expected of primary school students. Our results demonstrate a ``jagged frontier'' of capability. While some models achieve high accuracies on certain visual reasoning skills, their ability is unevenly distributed and improvements are most needed for ``dynamic'' visual reasoning such as \textit{folding, reflection, and rotation}.

\subsection{AI models demonstrate a ``jagged frontier'' where their low performance on visual reasoning has issues for supporting children's learning}

The VRB results highlight the jagged frontier of modern AI where models that achieve state-of-the-art results on textual reasoning benchmarks falter on primary-level visual logic. To support learning effectively, AI must meet a minimum capability threshold in visual reasoning. An AI tutor that can solve complex textual mathematics but then fails to recognise a student's error in a simple pattern-completion task is not yet trustworthy for unsupervised classroom use. Similarly, an MLLM for grading tasks that cannot provide accurate results or feedback on a subset of visually grounded questions risks reinforcing or even creating student misconceptions. A further challenge is the `invisible' nature of such failures, where trust has been built around a tool due to its strong performance across certain tasks, reducing the likelihood that teachers or learners will monitor outputs for hallucinations or subtle errors.

The lack of robustness in model accuracy on questions that contain minor errors typical of learning materials raises further concerns about classroom suitability, especially in settings where photocopying is the norm. This limitation may also erode potential time-saving benefits of AI if teachers must spend additional time improving material quality before use. Taken together, the VRB findings suggest that classroom applications involving visual reasoning must be implemented with explicit human oversight and quality-assurance checks.

\subsection{AI models show evidence of being constrained by ``the spatial ceiling'' and improvements are most needed on ``dynamic'' visual reasoning}

The failure patterns that we see in the VRB appear similar to the established cognitive frameworks in human spatial reasoning. This distinguishes between ``static'' or object-based skills and ``dynamic'' or spatial transformation skills \citep{newcombe2014}. In humans, tasks involving dynamic tasks such as mental rotations or recognising reflections tend to be more difficult to solve than those requiring static skills such as recognising structural features \citep{caissie2009,green1992}. For example, studies of children aged 9--12 have recorded accuracies between 73\% and 81\% for object-based skills while scores for spatial tasks involving visualisation and spatial relations dropped to between 55\% and 72\% \citep{soluki2021}.

The VRB reveals a similar hierarchy of difficulty for AI models. While contemporary models perform better on static skills such as\textit{ counting} and \textit{scaling}, their performance reaches a spatial ceiling when faced with dynamic operations such as \textit{folding, rotation}, and \textit{reflection}. This indicates that while models have acquired an object-level grasp of images, they lack the fundamental internal mechanisms to simulate spatial movements.

This finding aligns with the ``spatial gap'' found in vision language research \citep{wang2024spatial}, where vision is treated as a supporting cue and secondary to linguistic priors. This leads to a collapse in relational logic when a task requires the model to navigate spatial transformations even if the object is perceived correctly \citep{wang2024spatial,chen2026babyvision}.

Furthermore, the performance drop in mental manipulation tasks suggests a deeper architectural limitation. Current MLLMs seem to struggle to maintain structural integrity and geometric identity through transformations. This deficit is characterized in recent literature as a fundamental failure of ``spatial imagination'' \citep{carpenter1990}, confirming that even frontier models still fall short of the visual competencies that emerge in early human development \citep{chen2026babyvision}.

This architectural bottleneck is made worse by the cognitive load of rule induction \citep{carpenter1990}. In humans, performance is higher on simple pairs (analogies) than on complex matrix patterns, primarily because the former requires integrating fewer visual features \citep{soluki2021,carpenter1990}. MLLMs face a parallel bottleneck in which, if the initial visual encoding fails to spot the relational features, the model is effectively reasoning from an inaccurate version of the image \citep{chen2026babyvision}. Without a solid grasp of what it is seeing, the model's reasoning becomes decoupled from the visual ground truth and leads to misgrounded explanations that lack reasoning fidelity \citep{wang2024spatial}.

\subsection{Limitations and Future Directions}

Our study is bounded by several limitations, including a focus on high-stakes aptitude exams that represent the upper end of difficulty for 11-year-olds rather than routine early primary tasks that are the core of foundational numeracy. Additionally, the VRB utilises static, single-turn interactions which do not evaluate a model's ability to provide incremental hints or respond to student diagrams in a dialogue. There is also a fundamental difference in processing: humans solve these problems using internal imagery, whereas models operate through visually conditioned language generation, meaning a correct answer does not necessarily imply human-like reasoning.

Future research can take this further through process-aware evaluation to ensure that models demonstrate reasoning fidelity and avoid shortcuts. By drawing on work such as ConceptARC \citep{beger2025}, future work could also require models to generate natural language explanations to distinguish between correct intended abstractions and correct unintended rules \citep{beger2025}. This would allow the differentiation of false patterns that work for specific demonstrations but fail to generalise to broader classroom contexts. Additionally, this would also help determine if failures on classroom tasks stem from a lack of high-order logic or a perceptual bottleneck. Further research could mitigate these perceptual hurdles by evaluating the impact of external tool use by granting models access to Python libraries for precise geometric processing \citep{beger2025}.

Finally, moving toward multi-turn benchmarking will better simulate real AI tutor use cases and determine whether models can identify why a student's visual reasoning is flawed and provide a more effective corrective scaffold.

\section{Conclusion}

The Visual Reasoning Benchmark (VRB) evaluates multimodal LLMs on classroom-authentic, minimal-text visual reasoning items from primary assessments in Zambia and India. Across a broad set of proprietary and open models, accuracy ranges from near-chance to 78\%, showing that primary-level non-verbal reasoning remains unsolved for current systems.

Performance is defined by two consistent limits. Firstly, accuracy can swing sharply across task format, producing a ``jagged frontier'' where strong performance on linear pattern completion coexists with failures on process matching and 2D matrix completion. Secondly, skill effects reveal a clear spatial ceiling: controlling for task category, dataset, and model identity, one finds that static skills (e.g.\ \textit{counting, scaling}) have slightly higher accuracy, while \textit{folding, reflection}, and \textit{rotation} pose the greatest challenges.

In realistic LMIC classroom conditions, this brittleness is a deployment risk, and implementation of tasks that require visual reasoning should be done with human oversight.

\begin{ack}
\end{ack}

\bibliographystyle{plainnat}
\bibliography{references}

\begin{thebibliography}{27}
\providecommand{\natexlab}[1]{#1}
\providecommand{\url}[1]{\texttt{#1}}
\expandafter\ifx\csname urlstyle\endcsname\relax
  \providecommand{\doi}[1]{doi: #1}\else
  \providecommand{\doi}{doi: \begingroup \urlstyle{rm}\Url}\fi

\bibitem[Beger et~al.(2025)Beger, Yi, Fu, Moskvichev, Tsai, Rajamanickam, et~al.]{beger2025}
C.~Beger, R.~Yi, S.~Fu, A.~Moskvichev, S.~W. Tsai, S.~Rajamanickam, et~al.
\newblock Do {AI} models perform human-like abstract reasoning across modalities?
\newblock \emph{arXiv preprint arXiv:2510.02125v3}, 2025.

\bibitem[Broadbent et~al.(2018)Broadbent, Hayden, Tate, and von Kotze]{broadbent2018}
D.~Broadbent, C.~Hayden, R.~Tate, and L.~von Kotze, editors.
\newblock \emph{11+ Non-Verbal Reasoning For GL Assessment}.
\newblock Coordination Group Publications Ltd.\ (CGP), 2018.

\bibitem[Cai et~al.(2025)Cai, Yang, and Hu]{cai2025mmiq}
H.~Cai, Y.~Yang, and W.~Hu.
\newblock {MM-IQ}: Benchmarking human-like abstraction and reasoning in multimodal models.
\newblock \emph{arXiv preprint arXiv:2502.00698}, 2025.

\bibitem[Caissie et~al.(2009)Caissie, Vigneau, and Bors]{caissie2009}
A.~F. Caissie, F.~Vigneau, and D.~A. Bors.
\newblock What does the mental rotation test measure? {An} analysis of item difficulty and item characteristics.
\newblock 2009.

\bibitem[Carpenter et~al.(1990)Carpenter, Just, and Shell]{carpenter1990}
P.~A. Carpenter, M.~A. Just, and P.~Shell.
\newblock What one intelligence test measures: a theoretical account of the processing in the {Raven} {Progressive} {Matrices} {Test}.
\newblock \emph{Psychological Review}, 97\penalty0 (3):\penalty0 404, 1990.

\bibitem[Chen et~al.(2026)Chen, Xie, Liang, He, Zhao, Yang, et~al.]{chen2026babyvision}
L.~Chen, W.~Xie, Y.~Liang, H.~He, H.~Zhao, Z.~Yang, et~al.
\newblock {BabyVision}: Visual reasoning beyond language.
\newblock \emph{arXiv preprint arXiv:2601.06521}, 2026.

\bibitem[Chollet(2019)]{chollet2019}
F.~Chollet.
\newblock On the measure of intelligence.
\newblock \emph{arXiv preprint arXiv:1911.01547}, 2019.

\bibitem[Chollet et~al.(2025)Chollet, Knoop, Kamradt, Landers, and Pinkard]{chollet2025}
F.~Chollet, M.~Knoop, G.~Kamradt, B.~Landers, and H.~Pinkard.
\newblock {ARC-AGI-2}: A new challenge for frontier {AI} reasoning systems.
\newblock \emph{arXiv preprint arXiv:2505.11831}, 2025.

\bibitem[Cobbe et~al.(2021)Cobbe, Kosinski, and Bavarian]{cobbe2021gsm8k}
D.~Cobbe, C.~Kosinski, and M.~Bavarian.
\newblock Training verifiers to solve math word problems.
\newblock \emph{arXiv preprint arXiv:2110.14168}, 2021.

\bibitem[Green and Kluever(1992)]{green1992}
K.~E. Green and R.~C. Kluever.
\newblock Components of item difficulty of {Raven}'s matrices.
\newblock \emph{The Journal of General Psychology}, 119\penalty0 (2):\penalty0 189--199, 1992.

\bibitem[Hendrycks et~al.(2021{\natexlab{a}})Hendrycks, Burns, Basart, et~al.]{hendrycks2021mmlu}
D.~Hendrycks, C.~Burns, S.~Basart, et~al.
\newblock Measuring massive multitask language understanding.
\newblock In \emph{International Conference on Learning Representations (ICLR)}, 2021{\natexlab{a}}.

\bibitem[Hendrycks et~al.(2021{\natexlab{b}})Hendrycks, Burns, Zou, et~al.]{hendrycks2021math}
D.~Hendrycks, C.~Burns, C.~P. Zou, et~al.
\newblock Measuring mathematical problem solving with the {MATH} dataset.
\newblock In \emph{NeurIPS Datasets and Benchmarks}, 2021{\natexlab{b}}.

\bibitem[Jiang et~al.(2024)Jiang, Zhang, Sun, Sourati, Ahrabian, Ma, Ilievski, and Pujara]{jiang2024marvel}
Y.~Jiang, J.~Zhang, K.~Sun, Z.~Sourati, K.~Ahrabian, K.~Ma, F.~Ilievski, and J.~Pujara.
\newblock {MARVEL}: Multidimensional abstraction and reasoning through visual evaluation and learning.
\newblock \emph{arXiv preprint arXiv:2404.13591}, 2024.

\bibitem[Liu et~al.(2025)Liu, Chen, Zhang, and Wang]{liu2025verify}
Y.~Liu, Z.~Chen, T.~Zhang, and X.~Wang.
\newblock {VERIFY}: A benchmark of visual explanation and reasoning fidelity for {MLLMs}.
\newblock \emph{arXiv preprint arXiv:2503.11557}, 2025.

\bibitem[Lu et~al.(2024)Lu, Wang, Pan, et~al.]{lu2024mathv}
P.~Lu, K.~Wang, J.~Pan, et~al.
\newblock {MATH-V}: Measuring mathematical reasoning in diagram-augmented contexts.
\newblock \emph{arXiv preprint arXiv:2402.14804}, 2024.

\bibitem[Natsheh and Karsenty(2014)]{natsheh2014}
I.~Natsheh and R.~Karsenty.
\newblock Exploring the potential role of visual reasoning tasks among inexperienced solvers.
\newblock \emph{ZDM}, 46\penalty0 (1):\penalty0 109--122, 2014.

\bibitem[Newcombe and Shipley(2014)]{newcombe2014}
N.~S. Newcombe and T.~F. Shipley.
\newblock Thinking about spatial thinking: New typology, new assessments.
\newblock In \emph{Studying visual and spatial reasoning for design creativity}, pages 179--192. Springer Netherlands, 2014.

\bibitem[Park et~al.(2025)Park, Lee, and Cho]{park2025me2}
J.~Park, D.~Lee, and K.~Cho.
\newblock Explain with visual keypoints like a real teacher: Multimodal solution explanation with {ME2}.
\newblock \emph{arXiv preprint arXiv:2504.03197}, 2025.

\bibitem[Parkinson and Cutts(2025)]{parkinson2025}
J.~Parkinson and Q.~Cutts.
\newblock Improving primary school pupils' spatial skills leads to computational thinking gains.
\newblock In \emph{Proceedings of the 30th ACM Conference on Innovation and Technology in Computer Science Education V.~1}, pages 646--652, 2025.

\bibitem[Purcar et~al.(2024)Purcar, Bocos, Pop, Roman, Rad, Mara, Crisan, R\u{a}du\c{t}-Taciu, Mara, Todor, and Triff]{purcar2024}
A.~M. Purcar, M.~Bocos, A.~L. Pop, A.~Roman, D.~Rad, D.~Mara, C.~Crisan, R.~R\u{a}du\c{t}-Taciu, E.~L. Mara, I.~Todor, and D.~G. Triff.
\newblock The effect of visual reasoning on arithmetic word problem solving.
\newblock \emph{Education Sciences}, 14\penalty0 (3):\penalty0 278, 2024.

\bibitem[Raven(1936)]{raven1936}
J.~C. Raven.
\newblock Mental tests used in genetic studies: The performances of related individuals on tests mainly educative and mainly reproductive.
\newblock Master's thesis, University of London, 1936.

\bibitem[Soluki et~al.(2021)Soluki, Yazdani, Arjmandnia, Fathabadi, Hassanzadeh, and Nejati]{soluki2021}
S.~Soluki, S.~Yazdani, A.~A. Arjmandnia, J.~Fathabadi, S.~Hassanzadeh, and V.~Nejati.
\newblock Comprehensive assessment of spatial ability in children: A computerized tasks battery.
\newblock \emph{Advances in Cognitive Psychology}, 17\penalty0 (1):\penalty0 38, 2021.

\bibitem[Wang et~al.(2024)Wang, Ming, Shi, Vineet, Wang, Li, and Joshi]{wang2024spatial}
J.~Wang, Y.~Ming, Z.~Shi, V.~Vineet, X.~Wang, Y.~Li, and N.~Joshi.
\newblock Is a picture worth a thousand words? {Delving} into spatial reasoning for vision language models.
\newblock \emph{arXiv preprint arXiv:2406.14852}, 2024.

\bibitem[Wang et~al.(2025)Wang, Li, Yin, Ran, and Liu]{wang2025mvmath}
P.~Wang, Z.~Li, F.~Yin, D.~Ran, and C.-L. Liu.
\newblock {MV-MATH}: Evaluating multimodal math reasoning in multi-visual contexts.
\newblock In \emph{CVPR 2025}, 2025.

\bibitem[Xie et~al.(2023)Xie, Lu, Li, et~al.]{xie2023mathvista}
Y.~Xie, P.~Lu, Y.~Li, et~al.
\newblock {MathVista}: Evaluating mathematical reasoning of foundation models in visual contexts.
\newblock \emph{arXiv preprint arXiv:2310.02255}, 2023.

\bibitem[Xu et~al.(2025)Xu, Wang, Wang, et~al.]{xu2025visulogic}
W.~Xu, J.~Wang, W.~Wang, et~al.
\newblock {VisuLogic}: A benchmark for evaluating visual reasoning in multi-modal large language models.
\newblock \emph{arXiv preprint arXiv:2504.15279}, 2025.

\bibitem[Zhang et~al.(2024)Zhang, Liu, Wang, et~al.]{zhang2024mathverse}
R.~Zhang, J.~Liu, T.~Wang, et~al.
\newblock {MathVerse}: Does your multi-modal {LLM} truly see the diagrams?
\newblock \emph{arXiv preprint arXiv:2403.14624}, 2024.

\end{thebibliography}

\newpage
\appendix

\section{Definition of Skill Tags and Task Categories}
\label{app:definitions}

\begin{longtable}{p{3cm} p{9cm}}
\toprule
\textbf{Skill Tag} & \textbf{Definition} \\
\midrule
\endfirsthead

\toprule
\textbf{Skill Tag} & \textbf{Definition} \\
\midrule
\endhead

\bottomrule
\endfoot

\rowcolor{gray!15}
Shape and Form & The question involves shape (regular or irregular) as a key aspect and the ability to distinguish subtle differences in shape or form, including shape changes and shapes being added in different configurations. Where shape change is not the primary process, for example if it occurs as a result of reflection, no tag is given. \\[6pt]

Counting & The question requires counting the number of objects or parts of objects to determine the solution. \\[6pt]

\rowcolor{gray!15}
Shading and Line Type & The question involves a change in shading, fill, or line style, including type (curved/straight, etc.). Shading may be a complete fill with lines, dots or similar, or another pattern of objects contained in a shape. \\[6pt]

Rotation & The question involves an object or parts of an object which have been rotated. This is commonly by 45$^\circ$, 90$^\circ$, or 180$^\circ$ but could also be other turns. Where either reflection or rotation could occur, both should be tagged. \\[6pt]

\rowcolor{gray!15}
Reflection & The question involves an object or parts of an object which have been reflected. This could be horizontal, vertical, or diagonal symmetry. Where either reflection or rotation could occur, both should be tagged. \\[6pt]

Scaling & The question involves objects or parts of objects which increase or decrease in size. This may be over a sequence of shapes or just 2 shapes getting smaller or larger. \\[6pt]

\rowcolor{gray!15}
Folding & This question involves a shape being folded and punched or cut before the pattern on the unfolded shape is determined, or questions where folds are made in an asymmetric way. \\[6pt]

Reconfiguring Shapes & This question involves objects or parts of objects changing position, being brought together or separated. They may or may not be touching but should not overlap. \\[6pt]

\rowcolor{gray!15}
Layering & This question involves identifying shapes hidden within larger figures and visualising objects which are placed one on top of the other, or which were on top of each other and are taken apart. \\

\end{longtable}

\begin{longtable}{p{3.5cm} p{8.5cm}}
\toprule
\textbf{Task Category} & \textbf{Definition} \\
\midrule

\rowcolor{gray!15}
Reasoning by Analogy & The question involves a comparison where the relationship between the first set of images needs to be matched by the relationship between the second image set. \\[6pt]

Odd One Out & The question involves a group of images where most are identical and the different one must be selected, or where most images share a common feature and the one without this feature is identified. \\[6pt]

\rowcolor{gray!15}
Pattern Completion (Linear) & The question involves a sequence. The last image should be chosen to continue the series in the same way. \\[6pt]

Pattern Completion (2D) & The question involves a grid structure with one part missing and needs to be filled in to complete the grid following the existing pattern. \\[6pt]

\rowcolor{gray!15}
Match (Process) & The question involves a group of question figures which show a process. The answer is chosen to match the output of this process. \\[6pt]

Match (Figure) & The question involves a single problem figure and several answer figures. The answer figure is chosen which matches the problem figure in the way described in the question. \\
\bottomrule
\end{longtable}

\section{Definitions of Error Types and Severity}
\label{app:errors}

\begin{longtable}{p{4cm} p{8cm}}
\toprule
\textbf{Error Type} & \textbf{Definition} \\
\midrule

\rowcolor{gray!15}
Image error & This is an issue with the images that make up the question (for example, inconsistent shading or inaccurate sizing). \\[6pt]

Question presentation error & This is an issue with the text in the question or another issue with the production of the question, such as a mouse cursor overlaying an image. \\
\bottomrule
\end{longtable}

\begin{longtable}{p{2cm} p{10cm}}
\toprule
\textbf{Severity} & \textbf{Definition} \\
\midrule

\rowcolor{gray!15}
Minor & A small error which is not likely to affect a student's ability to understand and answer the question. The correct answer is clear. It is likely to be either easily recognisable (such as a misprint with a mouse cursor over the image), not essential for answering the problem (such as an absent dotted line for a fold), or does not have a significant impact on meaning (such as minor grammar issues). \\[6pt]

Medium & A more substantial error which could affect a student's ability to distinguish between the options, but the correct answer remains clear. For example, an issue with degrees of rotation where one correct answer still stands out or misleading wording in the written question where the visual image is still clear. \\
\bottomrule
\end{longtable}

\section{Additional Figures}
\label{app:additional}

\begin{figure}[H]
    \centering
    \includegraphics[width=\textwidth]{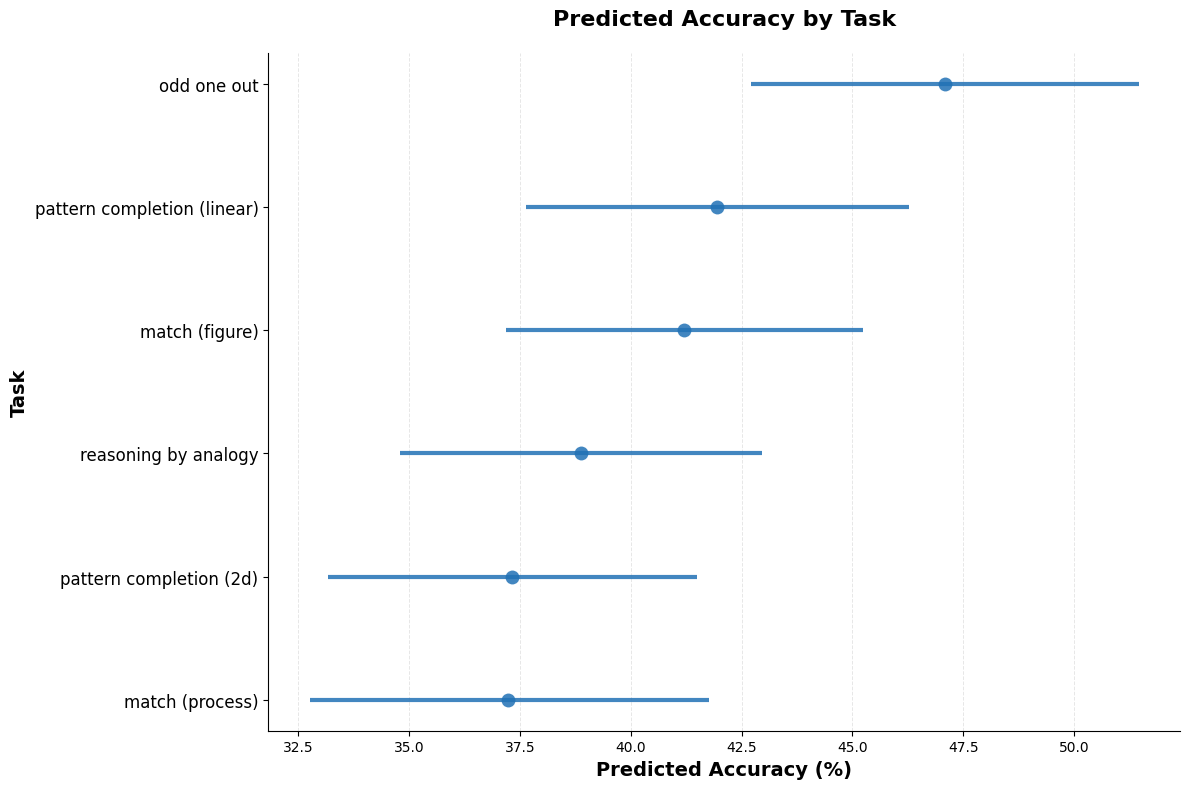}
    \caption{Task effects on item difficulty ranked from most difficult (bottom) to easiest (top).}
    \label{fig:task_difficulty}
\end{figure}

\begin{figure}[H]
    \centering
    \includegraphics[width=\textwidth]{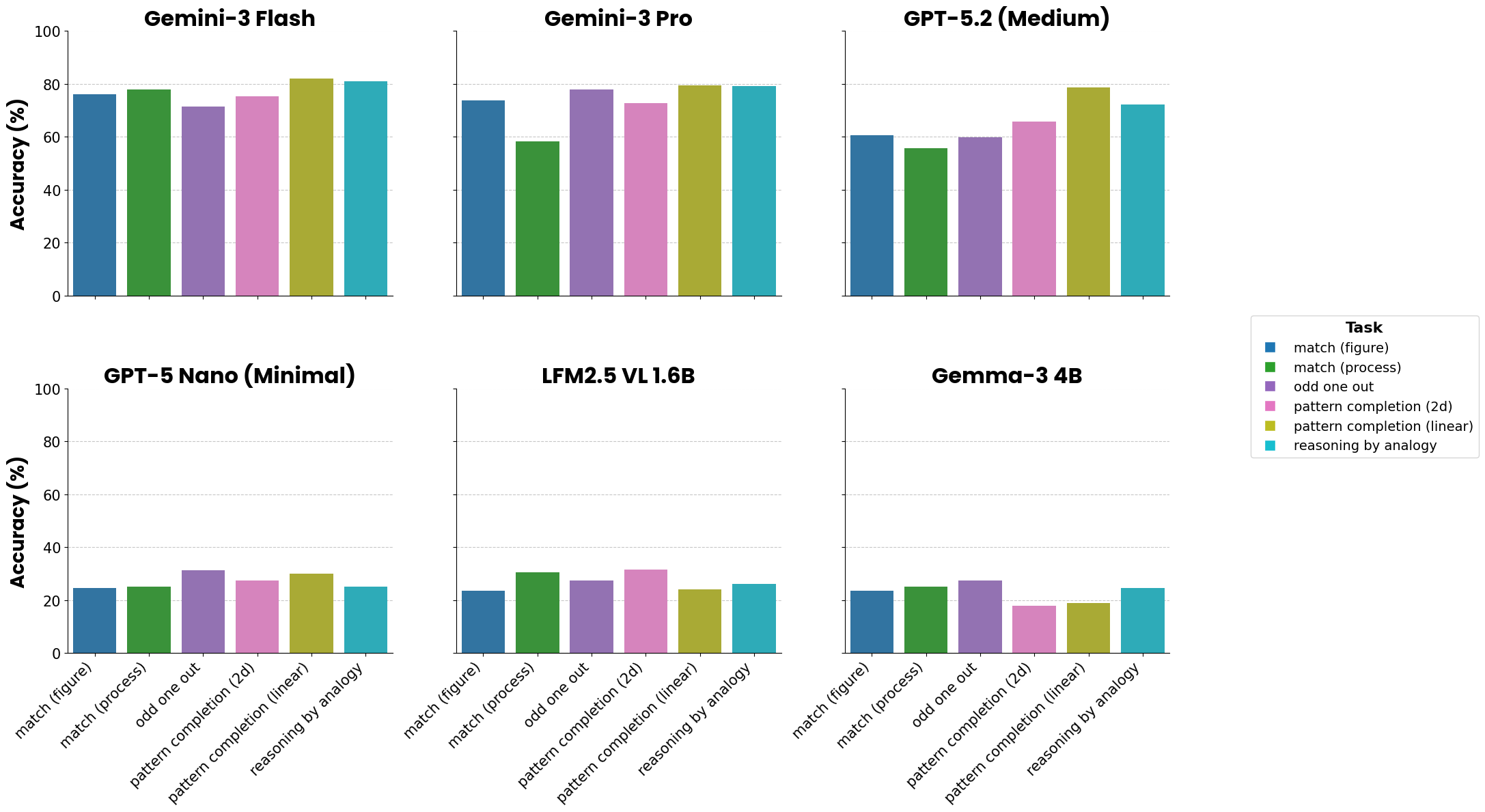}
    \caption{Per-model category bar charts (Top and bottom 3 models).}
    \label{fig:per_model}
\end{figure}

\section*{Extended Experimental Details}

\section*{Data / Code Availability}

The evaluation prompts, scoring code, and analysis pipeline are publicly available at \url{https://github.com/AI-for-Education/vrb-benchmark}. The question images and answer keys are held privately to preserve benchmark integrity; access can be arranged by contacting the authors.

\end{document}